%% file: acl2021.tex
\documentclass[11pt,a4paper]{article}
\usepackage[final]{emnlp2022}
\usepackage{times}
\usepackage{latexsym}

\usepackage{booktabs}
\usepackage{multicol,multirow}
\usepackage{graphicx}
\usepackage{todonotes}
\usepackage{amsmath}
\usepackage{xspace}
\usepackage{xfrac}

\usepackage{microtype}

\usepackage{comment}
\newcommand\modelnamenoxs{{\mbox{\fontfamily{qcr}\selectfont MaskEval}}}
\newcommand\modelname{\modelnamenoxs\xspace}

\usepackage{paralist}

\title{\protect\modelname: Weighted MLM-Based Evaluation  \\for Text Summarization and Simplification} 

\author{Yu Lu Liu$^{1,4}$ \quad  Rachel Bawden$^2$ \quad Thomas Scialom$^{3}$\thanks{~~Most of TS's contributions were made while he was a PhD student at Sorbonne Université.} \\ \textbf{Benoît Sagot}$^2$ \quad \textbf{Jackie Chi Kit Cheung}$^{1,4}$ \\

$^1$MILA, Montreal, Canada \quad
$^2$Inria, Paris, France \quad $^3$META AI, Paris, France \\
$^4$School of Computer Science, McGill University, Montreal, Canada \\

\texttt{yu-lu.liu@mila.quebec} \quad \texttt{firstname.lastname@inria.fr} \\
\texttt{tscialom@fb.com} \quad \texttt{jcheung@cs.mcgill.ca}
}

\begin{document}
\maketitle
\begin{abstract}
In text summarization and simplification, system outputs must be evaluated along multiple dimensions such as relevance, factual consistency, fluency, and grammaticality, and a wide range of possible outputs could be of high quality. These properties make the development of an adaptable, reference-less evaluation metric both necessary and challenging. We introduce \modelname, a reference-less metric for text summarization and simplification that operates by performing masked language modeling (MLM) on the concatenation of the candidate and the source texts. It features an attention-like weighting mechanism to modulate the relative importance of each MLM step, which crucially allows it to be adapted to evaluate different quality dimensions. We demonstrate its effectiveness on English summarization and simplification in terms of correlations with human judgments, and explore transfer scenarios between the two tasks.

\end{abstract}

\section{Introduction}

\begin{figure*}[!ht]
    \centering
    \includegraphics[width=0.9\textwidth]{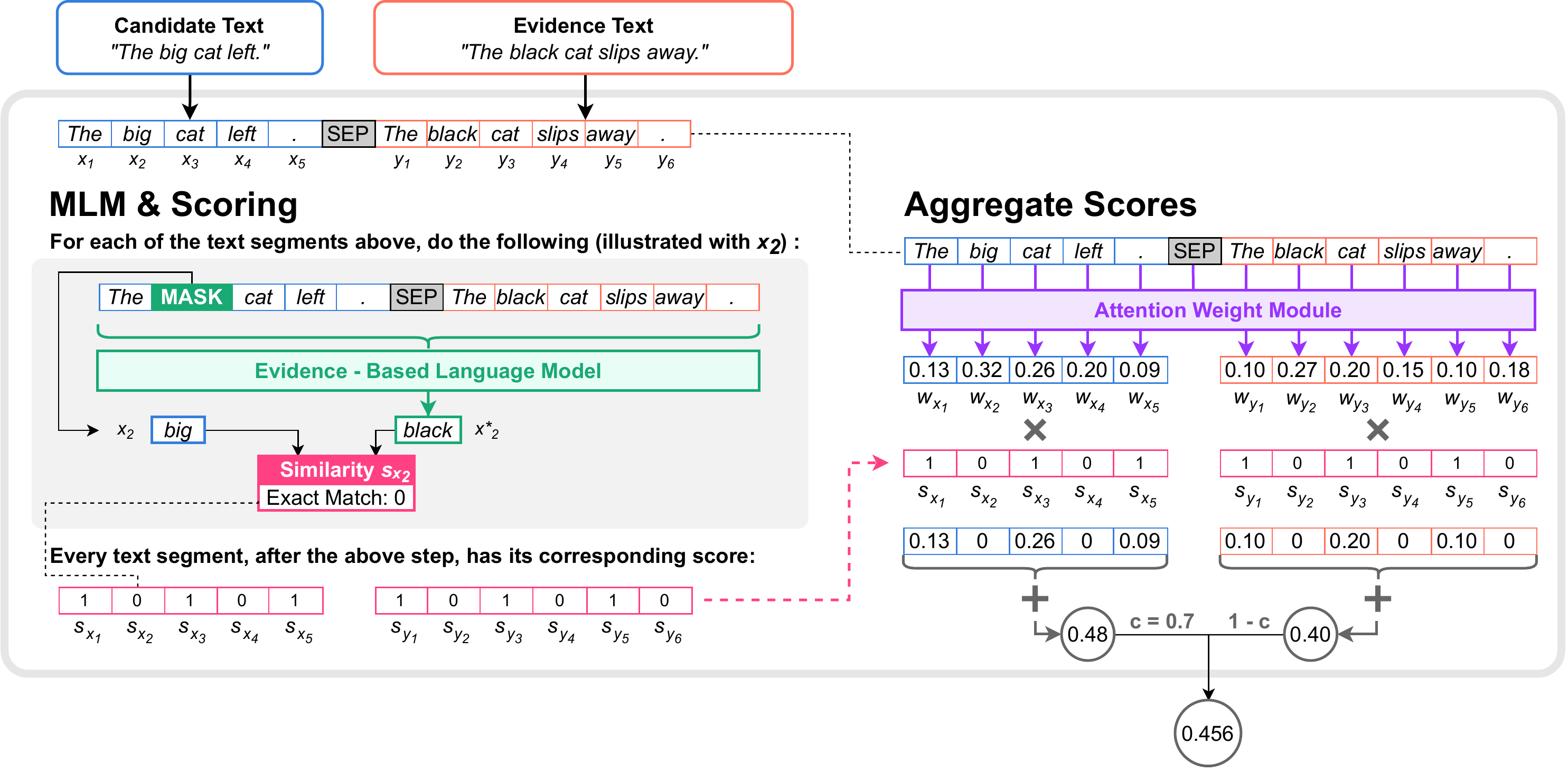}
    \caption{Illustration of the \modelname framework, which consists of two steps: (i)~masked language modeling (MLM) and (ii)~score aggregation. More details are provided in Section~\ref{sec:method}.}
    \label{fig:pipeline_example}
\end{figure*}

Automatic evaluation metrics are central to measuring progress in natural language generation (NLG) \citep{callison-burch-etal-2006-evaluating,graham-2015-evaluating,martin-etal-2018-reference}. Particularly challenging is the development of metrics for tasks such as summarization and text simplification.  Compared to machine translation (MT), where good outputs are limited to those that reproduce all input information, there is a wider range of good summaries/simplifications because the degree of succinctness/simplicity of the output can vary greatly. A further complication is that multiple qualities of the output text must be evaluated, such as factual consistency or fluency.

For such tasks, traditional reference-based metrics such as ROUGE \citep{lin-2004-rouge} and BLEU \citep{papineni-etal-2002-bleu} can therefore be limited by the diversity of the available references.\footnote{Multiple references, including automatically generated ones \cite{bawden-etal-2020-study}, can improve this scenario, but cannot cover all possibilities and are also costly to produce.} Indeed, previous work has shown their limited correlation with human quality judgments \citep{callison-burch-etal-2006-evaluating,novikova-etal-2017-need,sulem-etal-2018-bleu}. 

A promising alternative is reference-less metrics which score a candidate text given only the source text. One such approach makes use of neural language models (LM) \cite{devlin-etal-2019-bert,lewis-etal-2020-bart,BERTScore,sellam-etal-2020-bleurt,rei-etal-2020-comet,BARTScore}. For example, BARTScore \citep{BARTScore} uses a LM to auto-regressively score one text (e.g.~a candidate) given another (e.g.~the source or the reference). This provides the means to exploit the LM for the task for which it was trained. {}
A second approach consists of question-based metrics \citep{wang-etal-2020-asking,durmus2020feqa,QuestEval}, which carry out automatic question generation and answering (QG/QA) based on the candidate and the source text. Typically, answers are assumed to be nouns which are extracted from the text.


Both approaches have achieved state-of-the-art correlation scores with human judgments, depending on the specific dataset and dimension of evaluation. However, there has been limited prior work in either paradigm on adapting a reference-less evaluation metric to multiple evaluation dimensions, tasks, and languages.\footnote{We do not evaluate in the multilingual setting due to current lack of evaluation data.}

In this work, we propose \modelname, an adaptable reference-less LM-based metric which draws on the strengths of both approaches above. Like prior LM-based approaches, it can exploit in-domain data for fine-tuning. However, it shares a key assumption of question-based metrics that not all tokens should be equally important for evaluating the output. In fact, we propose to \emph{learn} this importance to further improve performance.

\modelname can be characterized by the following: (i)~it features a masked language modeling task (MLM) over both the candidate and source text, inspired by the translation modeling objective (TLM) \citep{lample-2019-cross}, and (ii)~learned weights that allow \modelname to vary the importance given to words. We use this second feature to analyze the contribution of certain classes of words, and to selectively mask inputs to reduce computational cost.

Our contributions can be summarized as follows:
\begin{compactitem}
    \item We introduce \modelname, a reference-less metric for text transformation tasks based on a modified MLM framework 
    and a novel learned weighter of words;\iftrue\footnote{Our code will be made publicly available at: \url{https://github.com/YuLuLiu/MaskEval}}\fi
    
    \item We evaluate \modelname on English summarization, surpassing the best previous question-based metric \citep{QuestEval} in three out of four dimensions, and the best previous LM-based metric \citep{BARTScore} in factual consistency and fluency;
    
    \item We show that \modelname trained on summarization data transfers well to simplification, and vice versa. We also show that weighters trained on summarization can improve the metric's performance on simplification. 
\end{compactitem}

\section{Related Work}\label{sec:related-work}

While $n$-gram-based reference-based metrics such as ROUGE \citep{lin-2004-rouge} and BLEU \citep{papineni-etal-2002-bleu} are perhaps the most established in NLG, two more recent approaches have been shown to provide better correlations with human judgments of quality while being reference-less: those based on pre-trained neural LMs and those based on QG/QA.

\paragraph{LM-based metrics}
Pretrained LMs have been used in different ways: (i)~by comparing aligned token-level embeddings between the candidate and reference text, as with BERTScore \cite{BERTScore}, (ii)~by fine-tuning them either to directly reproduce human quality judgments \cite{sellam-etal-2020-bleurt} or to rerank pairs of candidate texts \cite{rei-etal-2020-comet}, and (iii)~by exploiting text-to-text pretrained LMs to score the candidate and source texts, as with BARTScore \cite{BARTScore}, in a similar way to PRISM \cite{thompson-post-2020-automatic}, which relies on multilingual paraphrasing as opposed to an LM. 
BERTScore has been shown to be poorly adapted to summarization \cite{scialom2021beametrics}. 


%

Both BARTScore and PRISM formulate the evaluation task as text generation, where the score is based on the log probability of the candidate being auto-regressively generated given the source text. 
While BARTScore achieves good correlations with human judgments for English summarization, 
it has a few potential disadvantages with respect to the way in which one text is scored based on the other: (i)~the model is auto-regressive, and therefore, while the text being scored is conditioned on the entirety of the other text, it is only conditioned on the left context of itself and (ii)~it uses a uniform weighting scheme, assigning an equal importance to each generation step (alternative weighting schemes were reported to give lower results). 

We seek to solve both of these disadvantages with \modelname, by (i)~replacing auto-regressive generation with successive masked language modeling (MLM), with prediction conditioned on both the candidate text and the source text, inspired by the translation language modeling (TLM) objective of XLM \citep{lample-2019-cross} 
and (ii)~learning a weighter to attribute varying importance to different words in the texts.

\paragraph{Question-based metrics}
A parallel direction is the development of question-based metrics \citep{matan2019,scialom-etal-2019-answers,durmus2020feqa,wang-etal-2020-asking,QuestEval}, where the idea is to automatically generate and then answer questions based on the candidate and the source text. The answers to the questions are nouns extracted from the texts. 
Different versions exist depending on which texts the questions/answers are conditioned on: 
\citet{scialom-etal-2019-answers} generate questions by using the source document 
while \citet{wang-etal-2020-asking} and \citet{durmus2020feqa} do so using the candidate text. 
QuestEval unified both approaches, enabling further improvement. 
Most of the proposed metrics based on question-answering have targeted summarization. 
%

One of the major advantages of question-based metrics is their interpretability, producing human-readable questions and answers, which can offer insights into how a candidate text is either good or bad. However, they are limited by the necessity to have good systems for question generation and answering. This requires large-scale and high-quality data, which are not available in many languages other than English \citep{riabi-etal-2021-synthetic}.

\section{\modelname Framework}\label{sec:method}


\modelname scores a candidate text with respect to its source text by weighting word-level scores (from both the candidate and source text) 
in a two-step process (illustrated in Figure~\ref{fig:pipeline_example}).

\begin{enumerate}
    \item \textbf{Successive MLM}: We perform successive MLM on the words of both the candidate and source text, comparing each prediction to the ground truth to produce one score per word.
    \item \textbf{Weighted Score Aggregation}: We aggregate the scores using a learned weighter (optimized to different quality dimensions) in order to vary the importance given to each word. 
\end{enumerate}

\subsection{Word-level Segmentation in MLM}
\label{sec:text_seg_section}

\begin{figure}[!ht]
    \centering
    \includegraphics[width=0.9\linewidth]{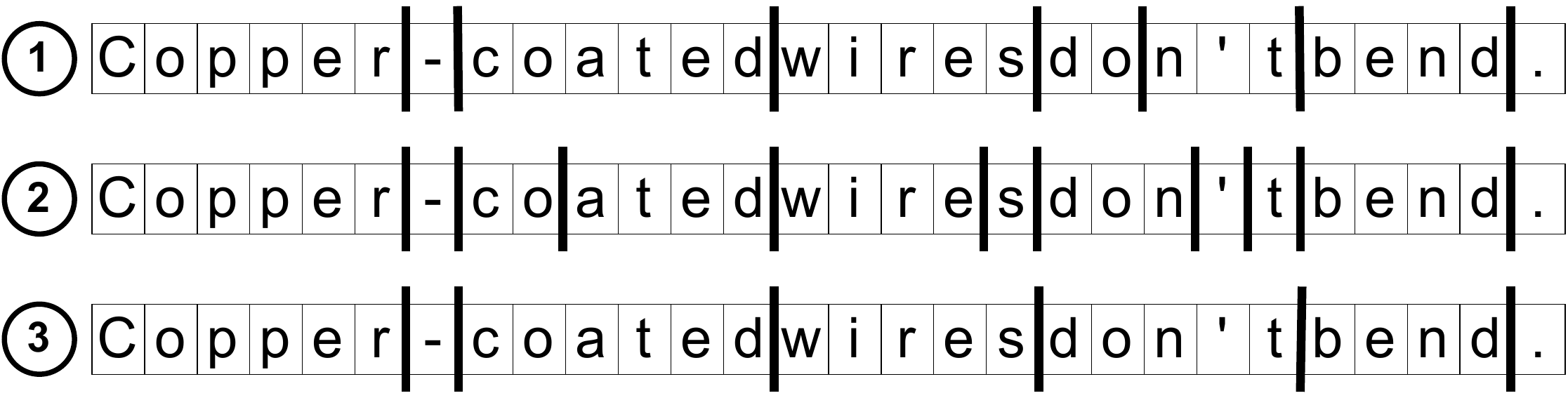}
    \caption{The proposed text segmentation (3) is the intersection of the boundaries between (1)~linguistically motivated tokenization produced by spaCy \citep{spacy2} and (2)~learned subword tokenization produced by WordPiece \citep{WordPiece}.}
    \label{fig:segmentation_example}
\end{figure}

In both steps, we choose to assign scores (respectively weights) to linguistically meaningful tokens (words as defined by a language-specific word-level tokenizer,\footnote{
We use spaCy tokenisers \cite{spacy2}.}) with the aim of making the method more interpretable and allowing us to perform linguistic analysis on learned weighters.
In order to ensure that word-level segmentation is consistent with the existing segmentation of the pretrained MLMs we use 
(i.e.~masking these linguistically defined units does not result in unnatural subword tokenization), 
%
we propose a method to reconcile the two by taking the intersection of their segmentation boundaries. An example of this method is shown in Figure~\ref{fig:segmentation_example}. We refer to each text segment resulting from this scheme as a ``word''. 

Given candidate text $x=(x_{1},\ldots,x_{N})$ containing $N$ words and source text $y=(y_{1},\ldots,y_{M})$ containing $M$ words, the aim is  to produce $N+M$ scores. For model-internal subword segmentation, we refer to the tokenized candidate text as $x=(t_x^{(1)},\ldots,t_x^{(n)})$ and the tokenized source text as $y=(t_y^{(1)},\ldots,t_y^{(m)})$, where $n$ and $m$ are the number of subword tokens in the candidate and source texts respectively ($N\leq n$ and $M\leq n$). We will use the notation $t_x^{(k)} \in x_i$ to represent the fact that token $t_x^{(k)}$ is part of word $x_i$.

\subsection{Masked Language Modeling}
The goal of this step is to produce a list of scores, each corresponding to an MLM step (i.e. a word). Intuitively, each score evaluates how well a trained model can predict the word when it is masked, given the other words in its text and the other text.



\paragraph{Masking and prediction} \label{masking} We first create a sequence by concatenating $x = (x_1, x_2, ..., x_N)$ and $y = (y_1, y_2, ..., y_M)$, placing a special separator mono-token word $\mathrm{\texttt{<sep>}}$ to denote the boundary between the two.
In this respect, our MLM resembles the translation language modeling objective introduced in XLM \citep{lample-2019-cross}.
Next, for each word position $i$ in the $x$ part of the sequence, we replace it with the mask token, thus creating masked sequence $m_{x_i}$. We take the original word $x_i$ as the ground-truth corresponding to this masked sequence. We do the same with each word position $j$ in the $y$ part of the sequence, resulting in masked sequence $m_{y_j}$ with ground-truth $y_j$. This results in $N + M$ masked sequences, each paired with its ground-truth word. The masked sequences are inputs to our MLM. 
We predict the masked words in the masked sequences $m_{x_i}$'s and $m_{y_j}$'s, denoting the predictions as:
\begin{align}
\hat{x}_i &= \textrm{MLM}(m_{x_i})\\
\hat{y}_j &= \textrm{MLM}(m_{y_j})
\end{align}

\paragraph{Scoring}
We score predictions $\hat{x}_i$ and $\hat{y}_j$ by computing their exact-match score against their corresponding ground-truth words $x_i$ and $y_j$.\footnote{We  considered other scoring functions: i)~computing the BERTScore between the prediction and the ground truth; ii)~the perplexity score of the predicted word, and iii)~the perplexity score of the ground-truth word. These scoring functions result in slightly worse performance than exact-match.} We give the score of 1 if the prediction and the ground-truth word are exactly the same and 0 otherwise.\footnote{
Both the prediction and the ground-truth word are lowercased before making the comparison.}
We denote the scores by:
\begin{align}
s_{x_i} = \textrm{Exact-Match}(x_i, \hat{x}_i)\\
s_{y_j} = \textrm{Exact-Match}(y_j, \hat{y}_j)
\end{align}

\paragraph{MLM Training} \label{MLM_training_method}
When fine-tuning our MLM (we fine-tune pre-trained MLMs), we create examples using the above procedure on existing datasets of document pairs. The only difference is that for each pair, we first randomly choose which text to mask (candidate or source text), and then randomly select one word position within the chosen document. 
This creates one masked sequence per pair of texts. We train using cross-entropy loss between the predicted word and the ground-truth word.

\subsection{Aggregation of Scores}\label{sec:weighter_description}
In order to produce a single quality score (which can be adjusted for different dimensions), we aggregate the scores from the previous step by computing a weighted sum as follows:
\resizebox{\linewidth}{!}{
\begin{minipage}{1.2\linewidth}
\begin{align}
\label{MaskEval_eqn}
\textrm{\modelname}(x, y) &= c \sum_{i=1}^{N} w_{x_i} s_{x_i}
+ (1-c) \sum_{j=1}^{M} w_{y_j} s_{y_j},
\end{align}
\end{minipage}}
\hphantom{add space otherwise too tight}

\noindent where $w_{x_i}$ (resp.~$w_{y_j}$) denotes the weight attributed to each word $x_i$ (resp.~$y_j$) of the candidate text $x$ (resp.~$y$), learned as described below and normalized such that $\sum^N_{i=1} w_{x_i}=1$ and $\sum^M_{j=1} w_{y_j}=1$.\footnote{Since $x$ and $y$ can be of different lengths, weights are defined and normalized separately for each text to avoid the longer text having more impact in the final score.}
$c$ denotes the learned weight (between 0 and 1) attributed to the candidate text.
The final \modelname score is between 0 and 1.

\begin{figure}[!ht]
    \centering
    \includegraphics[scale=0.535]{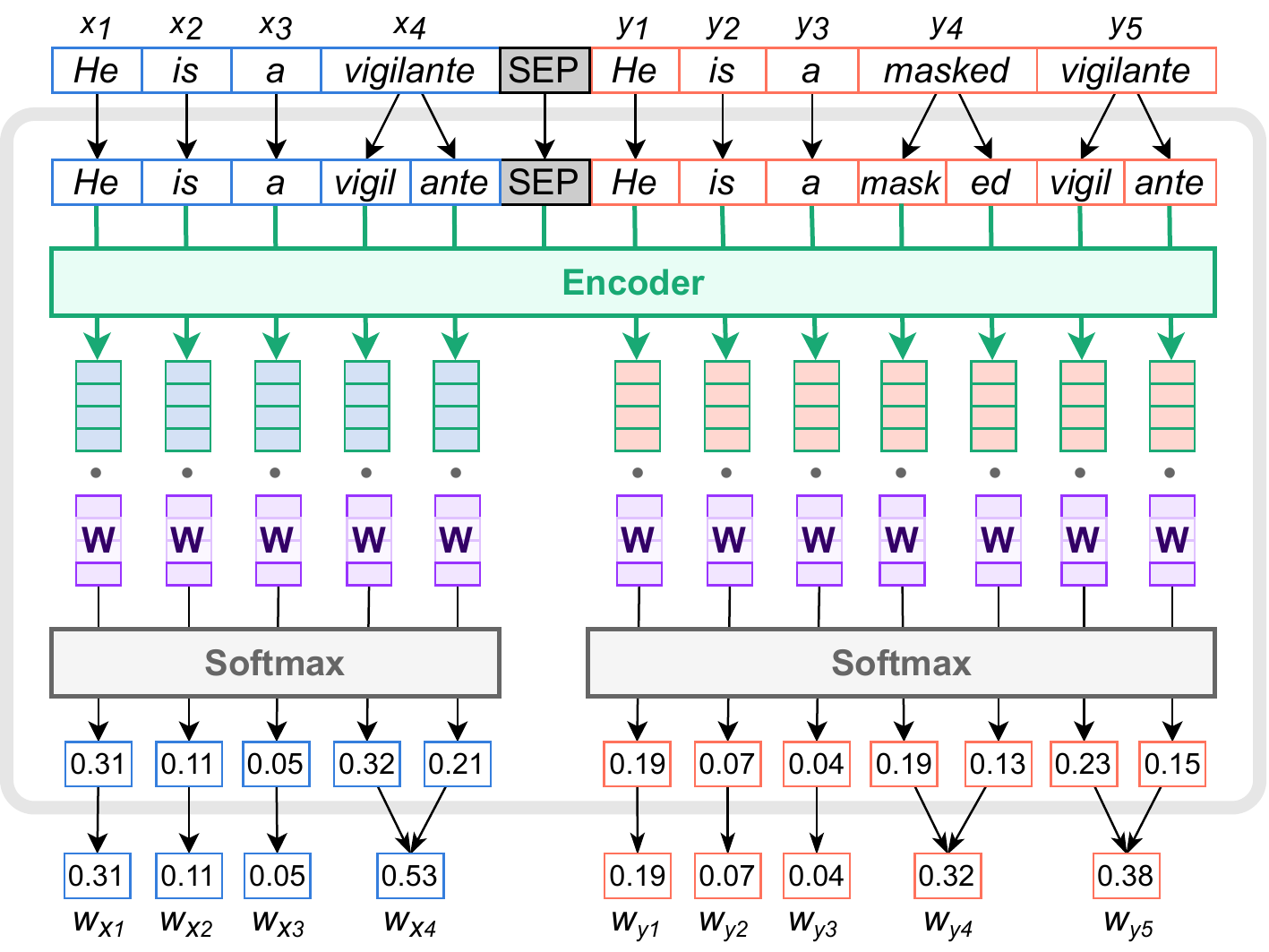}
    \caption{\modelname's word weighter. We (i)~extract contextual embeddings of the candidate and source texts, (ii)~apply a linear layer $W$ over the token embeddings, (iii)~apply a softmax over tokens for each text 
    and (iv)~regroup token scores to form word scores (a word's weight is the sum of its tokens' weights).} 
    \label{fig:Attention_Weighter}
\end{figure}

\subsubsection{Learned weights}
The weights $w_x$ and $w_y$ are learned using a separate attention-like module such that greater attention weights to words of interest, varying the importance given to each MLM step depending on their utility for the final score.  

As for the MLM step (Section~\ref{masking}), the input to the weighter is the concatenation of the tokenized candidate and source texts: $\left(t_x^{(1)},\ldots,t_x^{(n)},\mathrm{\texttt{<sep>}}, t_y^{(1)},\ldots,t_y^{(m)}\right)$.
We encode this input with a pretrained language model, resulting in contextual embeddings $e_x^{(k)}$ (resp.~$e_y^{(k')}$) for each token in $x$ (resp.~$y$). We then apply an attention-like mechanism over these embeddings, in the form of a linear layer $W$ followed by a separate softmax function for $x$ and for $y$, such that the scores for each sums to one. This can be expressed as follows (shown here for $x$, with $1\leq k\leq n$):
\begin{align}
\label{eqn: attn_layer}
v_k &= \sigma\!\left(W \cdot \left(e_x^{(1)}\ldots\  e_x^{(n)}\right)\right)_k
\end{align}

The weight $w_{x_i}$ attributed to each word $x_i$ is finally computed as the sum of its tokens' weights:
\begin{align}
\label{weights_eqn}
w_{x_i} = \sum_{k \mathrm{~s.th.~} t_x^{(k)} \in x_i}  v_k
\end{align}
The same process is applied to $y$, resulting in $w_{y_j}$ for each word $y_j$ in $y$. An illustration of the weighter is shown in Figure~\ref{fig:Attention_Weighter}.

\paragraph{Weighter Training}
The weighter is a regressor model  trained to mimic human judgment scores and therefore can be adapted to evaluate candidate texts along different dimensions of interest (e.g.~factual consistency), as long as data annotated for those dimensions exists.
Given a candidate paired with a source or reference text, and a human evaluation score for the candidate in the dimension we are optimizing for, we compute the \modelname score (Equation~\ref{MaskEval_eqn}) using the weights in Equation~\ref{weights_eqn}. The weighter's loss function is the mean squared error between the \modelname score and the human evaluation score.
\paragraph{Baseline weighting schemes}
As a baseline, we consider \modelname with uniform weights, a setup where $w_{x_i} = \sfrac{1}{N}$ for all words $x_i$ in $x$, $w_{y_j} = \sfrac{1}{M}$ for all words $y_j$ in $y$, and $c = \sfrac{1}{2}$. Since some quality of a candidate text (e.g. its fluency) should not be influenced by the source text, We also consider candidate-only \modelname, a setup where $w_{x_i} = \sfrac{1}{N}$ for all words $x_i$ in $x$, and $c = 1$. 

\section{Experiments}\label{sec:experiments}

\subsection{Experimental Details}

We  evaluate \modelname on English summarization and simplification. 
See Appendix~\ref{sec:appendix} for additional training details. 

\paragraph{MLMs}
We train two MLMs: one for summarization and one for simplification. Both are initialized with the T5 base model \cite{t5}, and then fine-tuned using the data  described in Section~\ref{sec:data}, following the process described in Section~\ref{sec:method}. To be consistent with T5’s 
training, we continue to use their masking format: the masked word is replaced with the token \texttt{$<$extra\_id\_0$>$}, and when fine-tuning we format the ground-truth output by placing it between the tokens \texttt{$<$extra\_id\_0$>$} and \texttt{$<$extra\_id\_1$>$}.

To keep the MLM steps reasonably memory-efficient, we use a maximum sequence length of 512 tokens. Each sequence, at both training time and inference time, is modified as follows: a sliding window is applied on the text being masked so that each masked token has a maximum of 24 tokens on each side. Then, the other text is truncated at the token level to respect the sequence length limit.
\paragraph{Attention Weight Module}  \label{sec:attn_weight_details}
The weighter takes as input contextual embeddings from the T5 base model. We train two sets of weighters using human-annotated data described in Section~\ref{sec:data}: one for summarization and one for simplification. The data has annotations of quality across different dimensions, and we train a weighter for each quality dimension, using the average score given by human annotators in said dimension as the ground-truth value (scaled to range from 0 to 1).

\subsection{Data}\label{sec:data}

\paragraph{Summarization}
Our MLM for summarization (\modelnamenoxs$_{\text{Summ}}$) is trained on the train set of CNN/Dailymail \citep{hermann2015teaching, CNNDMb} ($\sim$287K documents and their summaries). 

To train the weighters for summarization and to evaluate our metric on summarization, we use SummEval \citep{DBLP:journals/corr/abs-2007-12626}, one of the largest human-annotated datasets for English summarization. The collection comprises 100 news articles, randomly selected from the test set of CNN/DailyMail \citep{hermann2015teaching}. It contains 1,600 summary-article pairs, each pair scored by three annotators with respect to four dimensions: consistency (\texttt{con}), 
coherence (\texttt{coh}), 
fluency (\texttt{flu}), 
and relevance (\texttt{rel}). 

We evaluate \modelnamenoxs$_{\text{Summ}}$ with uniform weights on the whole of SummEval, allowing us to compare its performance with 
existing metrics (listed in Section~\ref{sec:comparison-metrics}).
%
To train the weighters, 
we use a subset of SummEval (700 randomly selected examples), and then test its performance 
on 
the remaining 900 examples.
We also evaluate the model with uniform weights  
on this same test subset to enable a fair comparison. 

\paragraph{Simplification}
We train our MLM for simplification, \modelnamenoxs$_{\text{Simpl}}$,
on WikiLarge's train set \cite{wikilarge} ($\sim$296K simplifications).

To train the weighters and to evaluate our metric, we use human simplification judgments provided in ASSET \cite{system_gen_simplification_data}. This data is composed of randomly selected sentences from TurkCorpus \cite{TurkCorpus}, with simplifications generated automatically (162 examples).
Each simplification was scored with respect to three dimensions: fluency (\texttt{flu}), 
simplicity (\texttt{sim}) and  
meaning preservation (\texttt{mea}).

We evaluate \modelnamenoxs$_{\text{Simpl}}$ with uniform weights on the whole 
test set, allowing us to compare 
to existing metrics (listed in Section~\ref{sec:comparison-metrics}).
We train the weighters on a subset of ASSET (62 randomly selected examples), and then test on 
the remaining 100 examples.
We also evaluate the model with uniform weights 
on this same test subset to enable a fair comparison.

\subsection{Task Transfer}
To explore transfer between  summarization and simplification tasks, we evaluate \modelname trained for one task on the other task{}, both with uniform weights and with the set of weighters trained for {}the other task. {}

\subsection{Comparison to Existing Metrics}\label{sec:comparison-metrics}


\label{Baselines}
As baseline metrics, we first consider the length and the perplexity of the hypothesis summary, as they are reported to perform as well as some evaluation metrics \cite{spurious_corr}.

\paragraph{Reference-based} We also consider three reference-based metrics: ROUGE \citep{lin-2004-rouge} and BLEU \citep{papineni-etal-2002-bleu}, and BERTScore \citep{BERTScore}. They compare a hypothesis text to one or more manually produced ground-truth texts, contrarily to \modelname, which is reference-less. For simplification, we additionally report SARI \cite{SARI}, a standard $n$-gram-based metric standard 
simplification.  

\paragraph{QA-Based} We consider three QA-based metrics for summarization: 
SummaQA \citep{SummaQA}, QAGS \citep{LeeQACE}, and QuestEval \cite{QuestEval}. For simplification, we report QuestEval only since it is the only one, to the best of our knowledge, to have been adapted to evaluate simplification \cite{human_gen_simplification_data}.

\paragraph{LM-based}
We compare to the two LM-based metrics closest to ours, in their reference-less variants: PRISM \cite{thompson-post-2020-automatic} and BARTScore \cite{BARTScore}.\footnote{The performance of both metrics on SummEval (Pearson correlation) are computed using outputted scores made available by at \url{https://github.com/neulab/BARTScore/tree/main/SUM/SummEval}. We report BARTScore, with the BART model finetuned with CNN/Dailymail \cite{hermann2015teaching, CNNDMb} and with prompt-tuning.}
\input{tab/1-col-eng-summ}
\input{tab/sys_gen_simplifications}
\section{Results}\label{sec:results}
\subsection{Summarization}
We report the Pearson correlation between \modelname scores and human judgments on the SummEval dataset in Table~\ref{tab:english_summarization}. 

\modelnamenoxs$_{\text{Summ}}$ achieves good scores on average, the candidate-only variant surpassing QuestEval by 4.3 points, although remaining below BARTscore by 1.3 points. The slightly lower average score for \modelnamenoxs$_{\text{Summ}}$ than BARTscore is mainly due to the lower score for coherence, which could be explained by the use of an MLM rather than auto-regressive decoding. However, it performs better than all previous metrics on two out of four dimensions, outperforming both BARTScore and QuestEval for consistency and fluency.  
%
The dimension that benefits most from our approach with respect to the previous best score is fluency 
(45.9$>$33.1). 

With learned weights, \modelnamenoxs$_{\text{Summ}}$ is able to improve its performance in all four dimensions with respect to uniform weighting, with relevance being improved the most (+18\%). 

\input{fig/POS-analysis}

\subsection{Simplification}
We report the Pearson correlation between \modelname scores and human judgments on the simplification evaluation set in Table~\ref{tab:sys_gen_simplification}. Without considering the transfer scenario, The highest performing metric is the reference-based BERTScore, when it has access to 10 references. The best reference-less metric is BARTscore, although PRISM is best for meaning preservation.  
\modelnamenoxs$_{\text{Simpl}}$ with uniform weighting has good correlations, behind both BARTscore and PRISM, but outperforming QuestEval.
Given the very small number of examples that could be used to train the weighters, \modelnamenoxs$_{\text{Simpl}}$ with learned weights is unable to improve its performance in any of the dimensions in ASSET. We nevertheless report these figures for completeness.

\subsection{Transfer Between Tasks}
We also report scores for transfer between the tasks in Tables~\ref{tab:english_summarization} and~\ref{tab:sys_gen_simplification}. \modelname with uniform weights trained for the task performs similarly to its counterpart trained for the other task. This shows that transfer is possible between the tasks.

For the weighted versions of the metrics, \modelnamenoxs$_{\text{Simp}}$ does not provide improvements when transferring, which is not unexpected given its poor performance on simplification. However positive results can be seen when transferring from the weighted version of \modelnamenoxs$_{\text{Summ}}$ to simplification. In particular, with weights optimized for summarization fluency, \modelnamenoxs$_{\text{Summ}}$ obtains state-of-the-art result on fluency, and greatly improves simplificity. 
Whilst it is expected that optimizing for summarization fluency improved simplification fluency, the improvements in the simplicity dimension are more surprising, and show the judgments for fluency are easily influenced by other evaluation dimensions. However, these results do show the potential to be able to transfer across dimensions from different tasks, which could be interesting when there are few annotations available.

%



\section{Discussion}\label{sec:discussion}
We choose to analyze our highest performing set of weighters, those optimized for summarization dimensions. In the following sections, the learned weight of a word $x_i$ (resp. $y_j$) includes the candidate weight, thus equaling $c\ w_{x_i}$ (resp. $(1-c)\ w_{y_j}$).
\subsection{Analysis of the Weighting Function} 
Figure~\ref{fig:weight_distribution} shows the average weight distribution across parts of speech on summary-source pairs from the test set of SummEval. We can see that by using weights optimized for fluency, \modelname  primarily uses MLM steps masking adpositions, determinants, and other POS tags (i.e. conjunctions, numbers, etc.) in the summary. This is expected since the fluency of a summary does not involve the source document. This equally applies to simplification, which could explain the great performance we obtain during task transfer. 

Some weights behave in an unexpected way: the weights optimized for factual consistency gives more importance to determiners than to nouns, which goes against the assumption commonly held by existing question-based metrics that nouns contain the most salient information. 


\subsection{Sparsity: Towards Selective Masking}
An important factor for an automatic metric to be widely adopted is computational efficiency. This was one of the important concerns with question-based metrics. 
%
%
We propose to use our weighters to selectively mask the input texts, only calculating scores based on the highest weighted MLM steps, in order to reduce the number of masked predictions while best maintaining performance. 

\begin{figure}[ht!]
    \centering
    \includegraphics[width=\columnwidth]{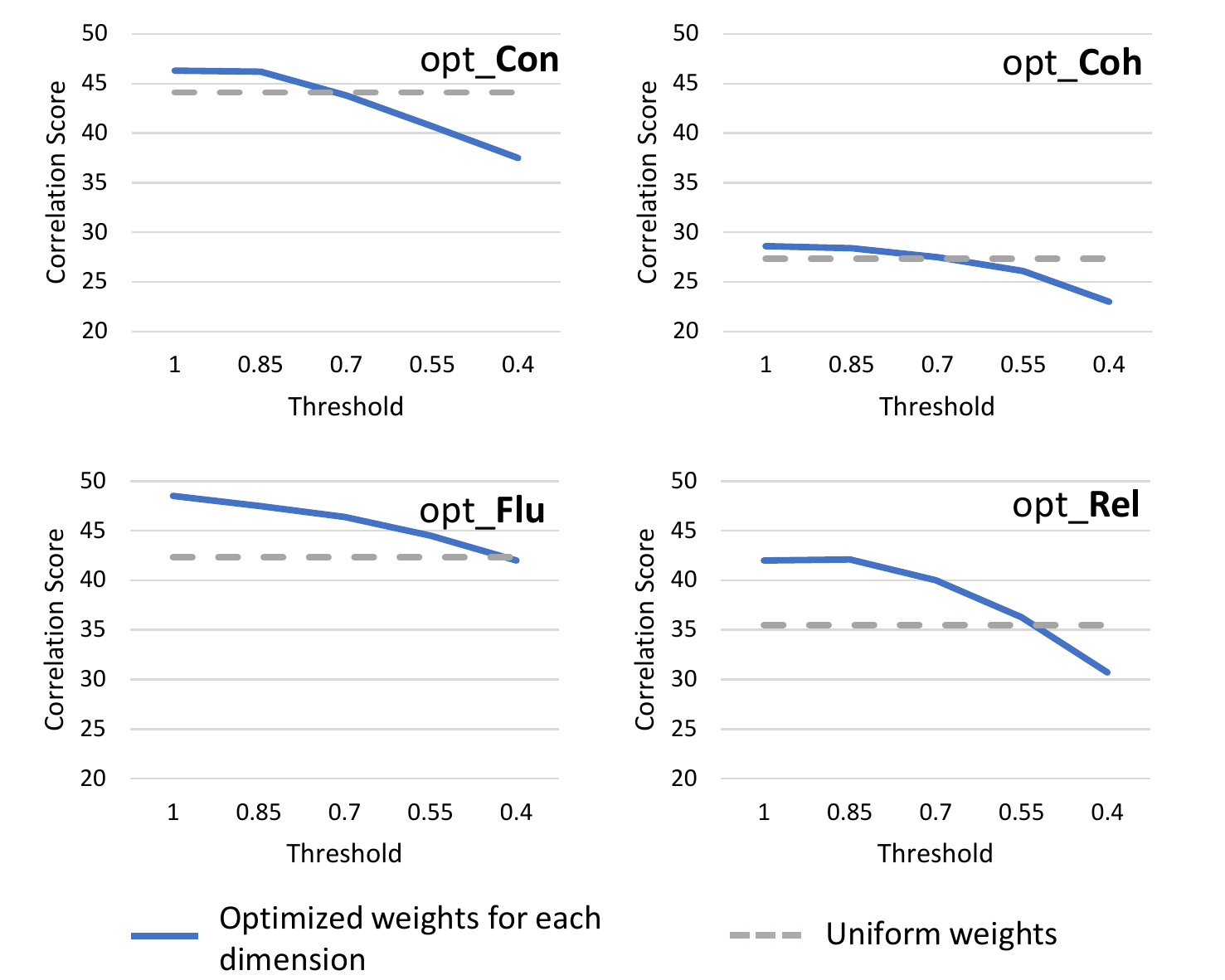}
    \caption{Correlations when optimizing for each dimension and computing the \modelname score with only the top MLM steps whose weights sum up to a set threshold.
    }
    \label{fig:sparsity_experiment}
\end{figure}

In Figure~\ref{fig:sparsity_experiment}, we report the Pearson correlation on the test subset of SummEval when only some weighted MLM steps are used in the computation of the \modelname score (i.e~we apply the weighter before the MLM step). We sort them by their weight , and only keep the top MLM steps whose learned weights sum up to a set threshold, computing the \modelname score using the retained MLM steps only. We can see that a threshold of $\sim$0.70 to preserve over 90\% of \modelname's original performance, and to match the performance of \modelname with uniform weights. This threshold  corresponds to only considering (on average) 25 to 130 MLM steps of a total of 480.
This suggests that with learned weights, the number of MLM steps necessary can be greatly reduced without compromising the performance. 

\section{Conclusion}
We have proposed \modelname, a metric for summarization and simplification that scores words in source and candidate texts using a MLM, then applies a learned weighting function over those scores, optimized to the task and evaluation dimension. 
Our analysis shows that different weights are applied depending on the dimension and that some parts of speech such as adpositions are more important than previously suspected. Our weighting function also allows us to produce a light-weight version of the metric, which uses only $\sim$20\% of the words to reach comparable correlation performance to using uniform weighting over all words.

In future work, we plan to test the approach for other text generation tasks (e.g.~MT) and for languages other than English, depending on the availability of such data.

\section*{Limitations}
Our LM-based metric can be easily adapted to a multilingual setting by finetuning a multilingual LM instead of the English one used in this paper. However, due to the lack of human annotations in multilingual summarization and simplification, we have not yet tested this capability. The capability of our metric to provide a good evaluation for texts from other tasks, other text generation systems, and other data distributions is also left to future work. Note that We considered using the Multi-SummEval dataset \citep{koto-etal-2021-evaluating} to test the multilingual capacity of our metric. However, we decided against this, due to potential problems we identified in the human annotation scheme employed in Multi-SummEval. Notably, (i)~annotators did not have access to source documents and annotated on the basis of a single reference text and (ii)~automatic evaluation metrics reported having a higher performance than human annotators. 

We measure the performance of our proposed metric by computing the correlation of its output scores with that given by human annotators. Our results are therefore reliant on the quality of evaluation datasets. For SummEval, for example, we have employed scores given by experts (annotators who have written academic papers on the topic of summarization). The expertise of these annotators may introduce a bias in the evaluation set as their judgments might differ from that of regular users of summarization systems.

\bibliographystyle{acl_natbib}
\bibliography{anthology,acl2021}

\appendix
\section{Implementation Details}
\label{sec:appendix}
\subsection{Training Details}
\paragraph{MLMs}
Both MLMs are fine-tuned for three epochs, with a batch size of 8 and the AdamW optimizer \citep{AdamW}, used with a constant learning rate of 1e-5. We trained using an NVIDIA QUADRO RTX 8000, with 48GB GPU memory. 20\% of the assigned training set is held-out during training to act as the validation set. We select the best checkpoint as being the one with the lowest validation loss.
\paragraph{Attention Weight Module}  \label{sec:appendix:attn_weight_details}
The weighter is trained over 100 epochs, using Adam optimizer \citep{Adam} with a constant learning rate of 1e-5. 20\% of the assigned training set is held-out during training to act as the validation set. Our final weighting function is produced by the checkpoint with the lowest validation loss. We randomly initialize weight vector $W$ (from Equation~\ref{eqn: attn_layer}) and initialize candidate weight $c$ (from Equation~\ref{MaskEval_eqn}) to $0.5$.

\end{document}

%% file: tab/1-col-eng-summ.tex
\begin{table}[!!ht]
\centering\small
\setlength{\tabcolsep}{4pt}
\begin{tabular}{lrrrrrr}
\toprule
\multicolumn{7}{c}{\bf{SummEval, with reference(s)}} \\
\midrule
                    & \#refs & \texttt{con} & \texttt{coh} & \texttt{flu} & \texttt{rel} & Ave. \\
\midrule
ROUGE-1             & 11    & 18.1        & 20.1      & 14.9    & 35.6      & 22.2    \\
ROUGE-L             & 11    & 15.7        & 15.6      & 13.8    & 33.4      & 19.6    \\
BLEU                & 11    & 17.5        & 22.0      & 13.7    & 35.6      & 22.2    \\
BERTScore-f         & 11    & 20.3        & 18.5      & 21.6    & 31.9      & 23.1    \\ \midrule
ROUGE-1             & 1     & 11.0        & 9.8       & 7.5     & 18.9      & 11.8    \\
ROUGE-L             & 1     & 8.2         & 7.3       & 5.7     & 13.5      & 8.7     \\
BLEU                & 1     & 8.9         & 3.9       & 4.0     & 12.7      & 7.4     \\
BERTScore-f         & 1     & 8.7         & 9.8       & 10.6    & 17.9      & 11.8    \\ 
\bottomrule
\end{tabular}
\begin{tabular}{lrrrrr} 

\toprule
\multicolumn{6}{c}{\bf{SummEval, reference-less}} \\
\midrule
                    & \texttt{con} & \texttt{coh} & \texttt{flu} & \texttt{rel} & Ave. \\
\midrule
Perplexity &  -3.1 & 15.7 & 8.9 & 19.8 & 10.3
\\
Length  & 8.1 & 8.6 & -2.9 & 26.6 & 10.1
  \\
\midrule
BARTScore           & 37.1        & \textbf{41.3}    & 33.1    & \textbf{44.8}    & \textbf{39.1} \\ 
PRISM                 & 29.7         & 28.1       & 24.8     & 29.7     & 28.1     \\
SummaQA               & 8.3         & 8.0       & -2.9    & 26.2      & 9.9     \\
QAGS                   & 20.4        & 7.7       & 16.8    & 9.1       & 13.7    \\
QuestEval &    42.0       & 24.0      & 28.4    & 39.2      & 33.5    \\ \midrule

\modelnamenoxs$_{\text{Summ}\ \text{uniform}}$  & 44.6       & 27.6      & 40.6   & 35.6     & 37.1    \\
\modelnamenoxs$_{\text{Summ}\ \text{candidate}}$ & \textbf{50.7} & 25.9 & \textbf{45.9} & 28.6 & 37.8 \\
\midrule
\modelnamenoxs$_{\text{Simpl}\ \text{uniform}}$  &  44.6 & 29.4 & 37.6 & 35.2 & 36.7 \\
\ \ \ \qquad\qquad\qquad$_{\text{opt\_\texttt{Flu}}}$ & 41.2 &  24.5 & 34.5 & 32.6 & 33.2  \\
\ \ \ \qquad\qquad\qquad$_{\text{opt\_\texttt{Sim}}}$ & 40.6 &  24.9 & 34.5 & 32.2 & 33.1  \\\ \ \ \qquad\qquad\qquad$_{\text{opt\_\texttt{Mea}}}$ & 41.5 &  25.2 & 34.8 & 32.3 & 33.5  \\

\midrule
\multicolumn{6}{c}{\em{SummEval subset (900 pairs)}} \\
\midrule
\modelnamenoxs$_{\text{Summ}\ \text{uniform}}$ & 44.1     & 27.3      & 42.3   & 35.5     & 37.3 \\
\modelnamenoxs$_{\text{Summ}\ \text{learned}}$ &46.3  & 28.6 & 48.5 &  42.0 & 41.4\\
\bottomrule
\end{tabular}
    \caption{English summarization results on the SummEval dataset (Pearson correlation). Unless indicated otherwise, the results are on the full SummEval test set. 
    Baseline non-QA metrics results are as reported in \citep{DBLP:journals/corr/abs-2007-12626}; QA-based metrics results are as reported in \citep{QuestEval}. 
    }
    \label{tab:english_summarization}
\end{table}

%% file: tab/sys_gen_simplifications.tex
\begin{table}[!ht]
\centering\small
\begin{tabular}{lrrrl}
\toprule
\multicolumn{5}{c}{\bf{ASSET, with reference(s)}}\\
\midrule
{} & \#refs & \texttt{flu} & \texttt{sim} & \texttt{mea} \\
\midrule
ROUGE-1 & 10 & 42.0 &	42.4 &	61.8  \\
ROUGE-L & 10 & 40.9 &	41.0 &	59.4 \\
BLEU & 10 & 28.9 &	29.5 &	47.6  \\
BERTScore-f & 10 & \textbf{58.0} & \textbf{54.7} & \textbf{73.4}\\
SARI & 10 & 34.4 &	29.9 &	51.9  \\
\midrule
ROUGE-1 & 1 & 33.7 & 31.2 & 47.9  \\
ROUGE-L & 1 & 31.8 &	28.5  & 43.0 \\
BLEU & 1 & 25.6 &	23.5 &	29.9 \\
BERTScore-f & 1 & 48.5 &	46.8 &	61.4  \\
SARI & 1 & 30.1 &	25.2 &	33.4  \\
\bottomrule
\end{tabular}
\begin{tabular}{lrrrl} 
\toprule
\multicolumn{4}{c}{\bf{ASSET, reference-less}} \\
\midrule
{} & \texttt{flu} & \texttt{sim} & \texttt{mea} \\
\midrule
Perplexity & 22.9 &	20.4&	23.1\\
Length & 2.5 &	-0.8 &	19.4 \\
\midrule
BARTScore & 57.5 &	\textbf{52.0} &	70.6 \\
PRISM & 56.8 &	45.1 &	\textbf{71.0}  \\
QuestEval & 33.9 &	32.7 &	63.4  \\
\midrule
\modelnamenoxs$_{\text{Simpl}\ \text{uniform}}$ & 50.5 & 43.6 & 67.5  \\
\modelnamenoxs$_{\text{Simpl}\ \text{candidate}}$ & 53.6 & 50.3 & 63.6  \\
\midrule
\modelnamenoxs$_{\text{Summ}\ \text{uniform}}$  &  48.6 & 
40.3 & 66.7 \\
$\ \ \ \qquad\qquad\qquad_{\text{opt\_\texttt{Con}}}$  & 49.8 & 42.8 & 61.0  \\
\ \ \ \qquad\qquad\qquad$_{\text{opt\_\texttt{Coh}}}$ & 39.6 & 31.1 & 61.9  \\
$\ \ \ \qquad\qquad\qquad_{\text{opt\_\texttt{Flu}}}$  & \textbf{58.7} & 51.8 & 56.9  \\
$\ \ \ \qquad\qquad\qquad_{\text{opt\_\texttt{Rel}}}$ & 44.0 & 34.1 & 65.0  \\
\midrule
\multicolumn{4}{c}{\em{Attention weight module's test set (100 pairs)}} \\
\midrule
\modelnamenoxs$_{\text{Simpl}\ \text{uniform}}$ &  49.5 & 46.3 & 68.8  \\
\modelnamenoxs$_{\text{Simpl}\ \text{learned}}$ & 39.7 & 32.9 & 58.7  \\
\bottomrule
\end{tabular}
    \caption{English simplification results on the ASSET dataset (Pearson correlation). Unless indicated otherwise, the results are on the full ASSET test set. 
    } 
    \label{tab:sys_gen_simplification}
\end{table}

%% file: fig/POS-analysis.tex
\begin{figure*}[!ht]
    \centering
    \includegraphics[scale=0.45]{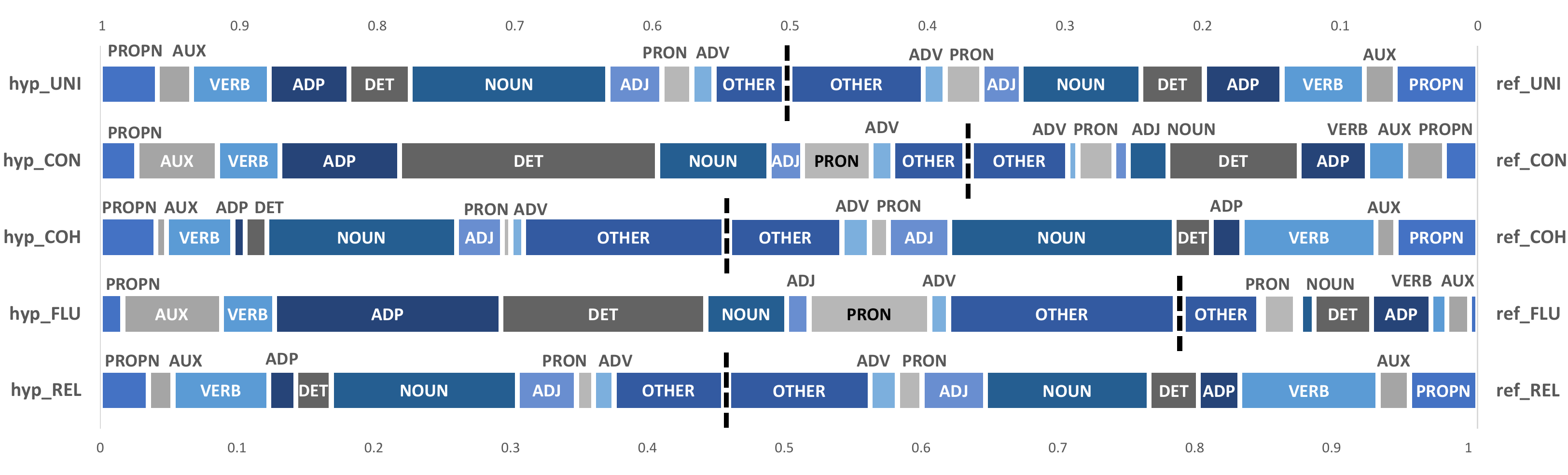}
    \caption{Average weight distribution across part-of-speech categories, as tagged by the English spaCy pipeline \citep{spacy2}, from the test subset of SummEval.}
    \label{fig:weight_distribution}
\end{figure*}